\documentclass[twocolumn]{article}

\usepackage{times}
\usepackage[dvips]{epsfig}
\usepackage{amsmath}
\usepackage{amssymb}

\begin{document}

\date{}

\title{DAMNED: A Distributed And Multithreaded Neural Event Driven
simulation framework}

\author{
Anthony {\sc Mouraud} \& Didier {\sc Puzenat}\\
GRIMAAG, Universit\'e Antilles-Guyane\\
Pointe-\`a-Pitre - Guadeloupe - France\\
{\small \{{\tt amouraud}, {\tt dpuzenat}\}{\tt @univ-ag.fr}}\\
\and
H\'el\`ene {\sc Paugam-Moisy}\\
Institute for Cognitive Science, UMR CNRS 5015\\
67 bld Pinel, F-69675 Bron cedex - Lyon - France\\
{\small {\tt hpaugam@isc.cnrs.fr}}\
}

\maketitle
\thispagestyle{empty}


\noindent
{\bf\normalsize ABSTRACT}\newline
In a Spiking Neural Networks (SNN), spike emissions are sparsely and irregularly distributed both in time and in the network architecture. Since a current feature of SNNs is a low average activity, efficient implementations of SNNs are usually based on an Event-Driven Simulation (EDS). On the other hand, simulations of large scale neural networks can take advantage of distributing the neurons on a set of processors (either workstation cluster or parallel computer). This article presents DAMNED, a large scale SNN simulation framework able to gather the benefits of EDS and parallel computing. Two levels of parallelism are combined: Distributed mapping of the neural topology, at the network level, and local multithreaded allocation of resources for simultaneous processing of events, at the neuron level. Based on the causality of events, a distributed solution is proposed for solving the complex problem of scheduling without synchronization barrier.

\vspace{2ex}

\noindent
{\bf\normalsize KEY WORDS}\newline
Spiking Neural Networks, Event-Driven Simulations, Parallel Computing, Multi-threading, Scheduling.


Accepted in : IASTED- PDCN 2006, International conference on Parallel and Distributed Computing and Networks

\section{Introduction}
\label{intro}

Advancing the knowledge on cognitive functions, simulations of Spiking Neural Networks (SNNs) represent a bridge between theoretical models and experimental measurements in neuroscience. Unlike usual threshold or sigmoid neurons, models of spiking neurons take into account the precise times of spike emissions. Therefore, SNNs help to simulate biologically plausible interactions between neurons and to study the influence of local parameters, at the neuron level, on the network global behavior, at the functional level. Results of very large scale SNN simulations can be analyzed the same way as experiments on animal or human, e.g. LFP \cite{Sin99} or EEG \cite{TBF01} recording, thus helping to understand how the brain works \cite{IGE04,MeuPau05}.
From a complementary point of view, theoretical studies \cite{Maa97,HopBro01} give large hope in the computational power of SNNs. Subject to the discovery of convenient learning rules \cite{LNM05,SimSga05}, simulations of large scale SNNs would provide efficient new solutions for many applications such as computer vision \cite{DGVT99}, adaptive control, real-time systems or autonomous robotics.

For developing very large scale SNNs, supporting a wide variety of spiking neuron models, a general purpose and fast running simulation framework is necessary. 
The well known GENESIS \cite{BB98} and NEURON \cite{HC97} are good for simulating precise biophysical models of neurons, but based on time driven simulation, i.e. scrolling all the neurons and synapses of the network at each time step, they are not specifically designed for fast simulation of very large scale SNNs. In accordance with biological observations, the neurons of an SNN are sparsely and irregularly connected in space (network topology), and the variability of spike flows implies they communicate irregularly in time (network dynamics) with a low average activity. Since the activity of an SNN can be fully described by emissions of dated spikes from pre-synaptic neurons towards post-synaptic neurons, an Event-Driven Simulation (EDS) is clearly suitable for sequential simulations of spiking neural networks \cite{Wat94,MDG00,Mak03,RocMar03,RGF03}. More generally, event-driven approaches substantially reduce the computational charge of simulators that control exchanges of dated events between event-driven cells EC, without checking each cell at each time step. On the other hand, since parallelism is an inherent feature of neural processing in brain, simulations of large scale neural networks could take advantage of parallel computing \cite{BAV99,EPPU02} or hardware implementations \cite{JSRMK97,Sei04,HGGMRK05}. A few studies coupled parallel computing and EDS, for general purpose systems \cite{Fer95}, and for SNN simulation \cite{GA98,PSWH01,GSW02}.

Our simulator belongs to the latter family and is close to Grassmann's work \cite{GA98,GSW02}, with some additional characteristics. Although it is known for long \cite{Pau95} that a fine grain mapping of the network (e.g. one neuron per processor) is dramatically inefficient, due to high communication overhead, we think that a multithreaded implementation of neurons as event-driven cells EC is efficient. Unlike several simulators \cite{Fer95,PSWH01,GSW02}, we avoid the implementation of a unique controller or farmer processor for scheduling the network simulation. We propose to direct the timing of execution through the times of events, without an explicit synchronization barrier. Hence we propose DAMNED, a ``Distributed And Multithreaded Neural Event Driven'' simulation framework that gathers the benefits of EDS and distributed computing, and combines two levels of parallelism (multiprocessor and multithread) for taking full advantage of the specific features of SNNs, whatever the models of spiking neurons to be implemented. Designed for efficient simulations either on workstation cluster or on parallel computer, the simulator is written in the object-oriented language C++, with the help of the MPI library to handle communications between distant processors. Section~\ref{event} develops the specificity of temporal events in SNNs. Section~\ref{archi} defines the distributed multiprocessor architecture and specifies the role of the multithreaded processes. Section~\ref{synch} details the algorithms and addresses the synchronization problem. In section~\ref{concl}, we conclude with an outlook on DAMNED exploitation.


\section{Temporal events in SNNs}
\label{event}

In a typical neural network, at any time, each neuron can receive on its dendritic tree some signals emitted by other neurons. An incoming signal arrives with a delay $d_{ij}$ and is weighted by a synaptic strength $w_{ij}$ to be processed by the soma. The values of $d_{ij}$ and $w_{ij}$ are specific to a given connection, from a presynaptic neuron $N_i$ to a postsynaptic neuron $N_j$. The membrane potential of a neuron varies in function of time and incoming signals. The neuron emits a spike, i.e. an outgoing signal on its axon, whenever its membrane potential overcomes a given threshold $\theta$. In experimental setting, and thus for simulations, firing times are measured with some resolution $\Delta t$, yielding a discrete time representation. Hence each spike can be considered as an event, with a time stamp, and each neuron can be considered as an event-driven cell $EC_j$, able to forecast its next spike emission time, as result from the integration of incoming spikes.

\begin{figure}[hbt]
\centerline{\epsfig{file=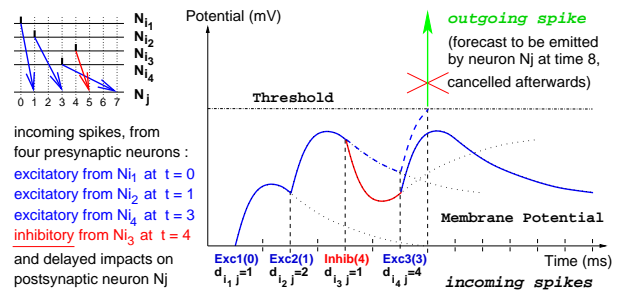, width=80mm}}
\caption{\small Variations of membrane potential for a postsynaptic neuron $N_j$. Three successive incoming Excitatory spikes let forecast an outgoing spike that must be cancelled afterwards, due to a further incoming Inhibitory spike, with a smaller delay $d_{i_3j}$.}
\label{potential}
\end{figure}

\vspace{-4mm}
However, the way to compute the future time of spike emission can be complex, depending on the model of neuron. For instance, if $EC_j$ manages the incoming delays, a further incoming spike, with inhibitory synapse, can cancel the forecast of an outgoing spike before the stamp associated to this event (see Figure~\ref{potential}). Hence we have to address the delayed firing problem (see \cite{GA98,Mak03,MDG00}).

Since the activity in the network is unpredictable, in order to preserve the temporal order of events, for the sake of biological plausibility, we ought to control the uncertainty of spike prediction. In the context of C++ language programming, we have chosen the following data structure for classes of ``events objects'' :
\begin{center}
CM event (resulting from {\em ComMunication})\\
= \ incoming spike, to be computed

{\small 
\begin{tabular}{|c|c|c|}
\hline
\vspace{-1mm}
label of & label of & time stamp of\\
 \ \ target neuron \ \  & \ source neuron \  & spike emission\\
$N_j$ (integer) & $N_i$ (integer) & $st_i$ (integer)\\
\hline
\end{tabular}
}

\bigskip

 CP event \ (resulting from {\em ComPutation})\\
= \ outgoing spike, to be emitted

{\small
\begin{tabular}{|c|c|c|}
\hline
\vspace{-1mm}
label of & time stamp of & \ \ \ certification \ \ \\
 \ source neuron \ \  & spike emission & flag\\
$N_i$ (integer) & $st_i$ (integer) & $crt$ (boolean)\\
\hline
\end{tabular}
}
\end{center}

\noindent where $crt$ is true only if the typical time of local run, on the processor implementing the neuron $N_i$, is high enough to guarantee that no further incoming spike could ever cancel the CP event (see section~\ref{synch} for further details).

Each class of ``event-driven cells'' EC objects is in charge of the computation methods associated to a model of spiking neuron, e.g. Integrate-and-Fire (IF, LIF), Spike Response Model (SRM), or other (see \cite{GK02,MB99}), so that different neuron types can be modeled in an heterogeneous SNN. Classically, an $EC_j$ object modeling a neuron $N_j$ has among its attributes the firing threshold $\theta_j$ of the neuron, the synaptic weights $w_{ij}$ and the delays $d_{ij}$ of the connections from all the presynaptic neurons $N_i$ able to emit a spike towards $N_j$.


\section{Distributed architecture and threads}
\label{archi}

\begin{figure*}[hbt]
\centerline{\epsfig{file=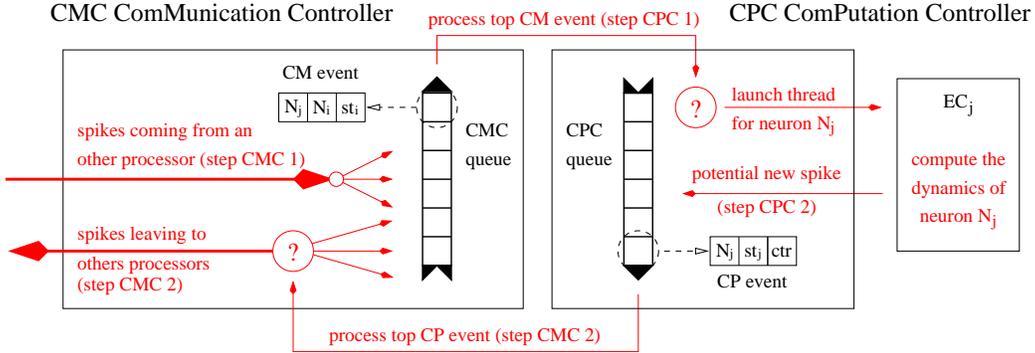,width=140mm}}
\caption{\small Architecture of a processor $Pr_p$, for $p \neq 0$. Several threads run simultaneously: A CMC thread, a CPC thread and as many threads as currently computing neurons. CMC receives spike events coming from other processors (step CMC 1). CMC inserts the incoming events in the CM priority queue according to their time stamp $st_i$. CPC checks if the top CM event is authorized for computation (step CPC 1). If authorisation is granted, the thread associated to $EC_j$ processes the [$N_j$,$N_i$,$st_i$] CM event. If $N_j$ triggers, the resulting spike generates a new CP events that are inserted in the CPC priority queue (step CPC 2). CMC checks if the top CP event is authorized for emission. If so, the spike event is dispatched towards all the target neurons, generating CM events that are inserted either in the CMC queue or in packets to be sent to other processors.}
\label{DAMNEDarchi}
\end{figure*}

The neural network topology must be distributed on $P$ processors according to a static mapping, to be defined as convenient for the application to be simulated. Each processor $Pr_p$ implements a certain amount of EC objects, labelled by their neuron number $N_i$. Each processor $Pr_p$ runs simultaneously two main threads so called CMC and CPC, for ``ComMunication Controller'' and ``ComPutation Controller'' respectively and as many extra threads as simultaneously computing neurons (see Figure~\ref{DAMNEDarchi}). Incoming spikes intended to be computed by every neurons $N_i$ belonging to processor $Pr_p$ are stored in a priority queue of CM events, ordered by their spike time stamp. Outgoing spikes resulting from computations of neurons $N_i$ belonging to processor $Pr_p$ are stored in a priority queue of CP events, ordered by their spike time stamp. They are intended to be sent by the CMC process to all the target neurons $N_j$ of $N_i$, whatever they belong to $Pr_p$ or to another processor. Processor $Pr_p$ knows the tables of postsynaptic neurons (neuron numbers $N_j$ and number $m$ of processor $Pr_m$ implementing $N_j$) for all its neurons $N_i$. For local target neurons, i.e. $N_j \in Pr_p$, a CP event $[N_i, st_i, crt]$ from the CPC queue generates CM events $[N_j, N_i, st_i]$ in the CMC queue of the same processor. For distant target neurons, each CM event $[N_k, N_i, st_i]$ is packeted into a message to be sent to processor $Pr_m$ implementing $N_k$.

As illustrated by Figure~\ref{DAMNEDarchi}, each processor runs in parallel: two main threads, CMC and CPC, with mutual exclusion for accessing each other priority queue. The CMC and CPC threads continuously run each an infinite loop, on the following procedures
\vspace{-2mm}
\paragraph{CMC, ComMunication Controller}
  \begin{enumerate}
  \vspace{-2mm}
  \item {\em message reception}: If messages from other processors are available, then place all the received CM events $[N_j, N_i, st_i]$ inside the CMC priority queue, ordered by their time stamp $st_i$ (or by arrival time, if equal time stamps exist),
  \vspace{-2mm}
  \item {\em emission control}: If the next outgoing spike $[N_i, st_i, crt]$, at the top of the CPC queue, is authorized, then look at the table of target neurons of $N_i$, create the CM events $[N_j, N_i, st_i]$ for all the postsynaptic neurons $N_j$ and place them either in the local CMC queue, if $N_j \in Pr_p$, or in packets prepared for further message sending,
  \vspace{-2mm}
  \item {\em message sending}: If packets are ready, then send them to the target processors.
  \end{enumerate}

All messages are sent and received according to MPI communication protocols. The {\em receive} and {\em send} procedures do not stall waiting for effective messages at each loop step. They only check if messages are present in their communication buffers. They process them if relevant, otherwise the loop goes on.
\vspace{-2mm}
\paragraph{CPC, ComPutation Controller}
  \begin{enumerate}
  \vspace{-2mm}
  \item {\em computation starter}: If the next incoming spike $[N_j, N_i, st_i]$ at the top of the CMC queue, is authorized, then launch the thread associated to $EC_j$ that implements neuron $N_j$
  \vspace{-2mm}
  \item {\em result collector}: If a new spike $[N_j, st_j, crt]$ has been generated by $EC_j$, then place the event inside the CPC priority queue, ordered by its time stamp $st_j$ (or default, arrival time if some other time stamps are equal)
  \end{enumerate}


Each time an incoming spike is computed by a neuron $N_j$, the associated thread is activated. Since the CPC runs an infinite loop on the {\em computation starter} and {\em result collector} procedures, several other threads, on EC objects, can be active simultaneously, thus implementing concurrent computation of several neurons, locally on processor $Pr_p$ (Figure~\ref{DAMNEDarchi}). On each processor, $nbth_p$ represents the number of active threads on $Pr_p$. The variable $nbth_p$ is incremented by the {\em computation starter} procedure each time a thread is activated for an EC object, and decremented by the {\em result collector} procedure each time a thread ends. The EC object keeps a pointer to every CP event it has generated as far as the event is present in the CPC queue. Hence the EC object can modify some certification flags if a new information allows it to authenticate some old-queued events that have not yet been emitted. For complete explanation of how to manage the delayed firing problem, let us define two other local variables:

\indent $et_p$ is the current {\em emission time} on processor $Pr_p$\\
\indent $pt_p$ is the current {\em processing time} on processor $Pr_p$\\

\noindent The variable $et_p$ switches to its opposite negative value if the CPC queue (spikes to be emitted) becomes empty, and switch back to opposite positive value when new events arrives in CPC queue. Same behavior for the variable $pt_p$ according to the state of the CMC queue (spikes to be processed). Those two variables play a fundamental role for controlling the scheduling on the set of processors and for defining the conditions of emission and computation authorizations, as detailed in section~\ref{synch}.

A last point about the distributed architecture: The neural network topology is spread onto $P$ processors. However, a realistic simulation requires an interaction with the environment. Hence an extra processor $Pr_0$ is necessary to send the environment stimuli to input neurons (distributed on several $Pr_p$, with $p \geq 1$) and to receive the response from output neurons (also distributed on several $Pr_q$, with $q \geq 1$). We prevent the processor $Pr_0$ to be a controller or a farmer, but the way it helps the scheduling of the whole simulation is also fundamental, as explained in next section.


\section{Synchronization control methods}
\label{synch}

The main point is to keep an exact computation of firing times for all the neurons of the network, and to prevent all the processors from deadlock situations, despite of the may-be irregularly distributed firing dynamics that can result from the environment data. The processor $Pr_0$ is in charge to send to the neural network all the input stimuli generated by the environment (e.g. translation in temporal coding of an input vector) and to receive all the spikes emitted by the output neurons of the network.

$Pr_0$ knows the actual time $T$ of the environment, and its current {\em emission time} $et_0$. At initial time, all the processors emission times $et_p$ are set to $0$. While the simulation runs, each processor, including $Pr_0$, may have a partial and obsolete view of the clocks of the other processors. Each processor $Pr_m$, $0 \leq m \leq P$, owns a local clock array $Clk(m)$ storing the emission times it currently knows, for all the processors $Pr_p$, $0 \leq p \leq P$

\medskip
$Clk(m)$ \ = \ \begin{tabular}{|c|c|c|c|}
\hline 
$et_{0}(m)$&
$et_{1}(m)$&
...&
$et_{P}(m)$\tabularnewline
\hline
\end{tabular}
\medskip

Each time a processor $Pr_p$ sends a packet of events (spike emissions) to a processor $Pr_m$, the message is encapsulated with the local clock $Clk(p)$. Hence the clock $Clk(m)$ can be updated each time $Pr_m$ receives a message. We assess that the whole network scheduling can be achieved this way, due to the local management of event causality. Since this way of controlling synchronization does not require ``look-ahead'' query-messages, we propose a more flexible method than the ``safe window'' solution described in \cite{GA98}.

\paragraph{Environment processor}

Processor $Pr_0$ is the only one that is not subject to the delayed firing problem since the environment processor relays all the input stimuli towards the neural network. Hence $Pr_0$ knows exactly the dates of all the external spikes that will trigger the input neurons of the SNN. Each time $Pr_0$ increments the actual time to $T$, all the packets with time stamps $T-1$ are ready for immediate sending. Messages are sent to all the processors $Pr_m$, $m \geq 1$, with the following information:
\begin{itemize}
\item the current update of the clock $Clk(0)$, where $et_0$ has just been incremented to $T$
\item if relevant, all the CM events $[N_j, Ext, T-1]$ that will generate spike emissions at time $T$ on $N_j$ input neurons owned by processor $Pr_m$
\end{itemize}
Even if a processor $Pr_m$ does not own input neurons, or if it owns input neurons that do not trigger at time $T$, it will receive a message with the clock $Clk(0)$. Note that the environment processor $Pr_0$ is the only one that can send messages reduced to the clock. Since all the processors are aware of the last update of $et_0$ ``immediately'', or as soon as the message can be transmitted [we assume reliable communication channels], the argument $(m)$ will be next omitted in notation $et_0(m)$.\\

The simulation starts running by the incrementation of $T$ to $1$. Since spike events communication must respect a causal order, the following conditions are always true, on every processor $Pr_m$ (arguments have been omitted):\\
{\small
\indent $T>0$ {\small and} $et_0 > 0$ {\small all along the run, after simulation start}\\
\indent $(\forall p \geq 1) \ \ et_0 \geq |et_p|$\\
\indent $T$ {\small and} $(\forall p \geq 0) \ \ |et_p|$ {\small are never decreasing}\\
\indent $(\forall p \geq 0) \ \ |pt_p|$ {\small is never decreasing}\\
\indent $(\forall p \geq 1) \ \ et_0 \geq |pt_p|$\\
}
\noindent The links between {\em emission time} $et_p$ and {\em processing time} $pt_p$ are clarified below, where algorithms that govern emission and computation authorizations are detailed.\\

Each time $Pr_0$ receives an output event packet, it forwards all the CM events $[Ext, N_i, st_i]$ to environment manager for further external processing, it updates its clock $Clk(0)$, from the received clock $Clk(q)$, as follows:\\
{\small
\indent $(\forall p \geq 1)$ if $|et_p(q)| \geq |et_p(0)|$ \ then $et_p(0) \leftarrow et_p(q)$;\\
\indent if $(\exists j \geq 1) \ |et_j(q)| = T$ \ then $T \leftarrow T+1$; $et_0 \leftarrow T$;\\
}
\noindent If $T$ has been incremented, then $Pr_0$ sends the appropriate messages to all the $Pr_m$. Note that $Pr_0$ may receive several output spike events, coming from different processors, between two successive increments of $T$. Conversely, it is possible that no output neuron send spike emission, at a given time $T$, and then $Pr_0$ does not receive any message and does not update its clock. Such a case would result quickly in stalling all the processors, by blocking their emission and computation authorizations. For preventing the system from deadlock, we assume that a time-out is running on $Pr_0$ and that $T$ is incremented when time-out expired, which is coherent with the notion of actual time represented by $T$. Hence $et_0$ is updated to $T$ and, provided that the time-out is sufficiently long, all the $et_p(0)$ can be set up to $T-1$. Messages are sent to all the processors, with updated clock $Clk(0)$ and possibly new spike events generated by external stimuli. This time-out is rarely activated but it prevents the system to fall into deadlock when the dynamics of the SNN is reduced to very low overall activity or activity loops that risk to be localized on a single processor or on a cluster with no output spikes.

\paragraph{CMC algorithms for emission authorization}
 $\ \ $ \\

On each processor $Pr_p$, for $p \geq 1$, the CMC runs an infinite loop on the successive procedures of message reception, {\em emission control} and message sending (see section~\ref{archi}).\\

At each message sending, processor $Pr_p$ checks if a packet of required size (minimal packet size $minpak$ is a parameter) is ready to be sent to another processor $Pr_m$. In case of successful checking, processor $Pr_p$ encapsulates the ready-to-be-sent packet with its current clock array $Clk(p)$ and sends it to the target processor.
At each message reception from a processor $Pr_q$, the CMC of $Pr_p$ inserts the incoming CM events in its priority queue, sets back the processing time to a positive value: {\small $pt_p \leftarrow |pt_p|$}, and updates its local knowledge of the $et_m$ on other processors as follows:\\
{\small
\quad $(\forall m \neq p)$ if $|et_m(q)| \geq |et_m(p)|$ \ then $et_m(p) \leftarrow et_m(q)$;\\
}

The {\em emission control} procedure picks up a new CP event $[N_i, st_i, crt]$ form the top of the CPC priority queue. This event has been computed by a local thread activated by the neuron $N_i$ and has generated a spike emission forecasted for time $st_i$. In order to respect the causality of events, the CMC process has to check the authorization to communicate this event, by the following algorithm:\\
{\small
if $st_i = et_p$ then emission is authorized;\\
\indent else if $crt$ then emission is authorized;\\
\indent \quad else if $st_i \leq pt_p$ then emission is authorized;\\
\indent \quad \quad else if $nbth_p = 0$ then\\
\indent \quad \quad \quad if $pt_p<0$ and $(\forall m\neq p) [st_i\leq et_m$ or $et_m\leq 0]$\\
\indent \quad \quad \quad \quad then emission is authorized;\\
\indent \quad \quad \quad \quad else emission is delayed;\\
\indent \quad \quad \quad else emission is delayed;\\
}
If the emission is authorized, the CMC process updates the local emission time: \quad {\small $et_p \leftarrow st_i$}\\
\noindent If the CP event is authorized then it is removed from the CPC queue. Each time the CPC queue becomes empty, the local emission time is changed to its opposite $et_p \leftarrow - et_p$ in order to indicate that there are no more spike emissions to communicate, at present time, on processor $Pr_p$. If the emission authorization generates, from the postsynaptic table of neuron $N_i$, new CM events to be further processed by one or more neurons local to $Pr_p$, then the processing time $pt_p$ takes back a positive value: {\small $pt_p \leftarrow |pt_p|$}.\\
 
The present algorithm controls that an authorization to be emitted can not be delivered to a spike event $[N_i, st_i, crt]$ before its validity has been assured, regarding to the overall run of the simulated SNN. The emission of the spike event is authorized if we are sure that all the further computations of neuron $N_i$ can not invalidate the present spike, either due to other computations locally running on processor $Pr_p$ (controls on $et_p$, $pt_p$ and $nbth_p$) or to distant spike events further incoming from other processors (controls on $et_m$, for all $m\neq p$). Even if a spike emission has been delayed only because the local clock $Clk(p)$ was not correctly updated, we avoid to overload the communication network with query messages, since the possible idle state is guaranteed to be ended by the reception of either new incoming events from other processors or clock messages coming from $Pr_0$. 


\paragraph{CPC algorithms for computation authorization}

On each processor $Pr_p$, the CPC runs an infinite loop on the successive procedures {\em computation starter} and {\em result collector} (see section~\ref{archi}).\\

The {\em computation starter} procedure picks up the top CM event $[N_j, N_i, st_i]$ of the CMC process priority queue. This event notifies that the neuron $N_i$ has emitted a spike at time $st_i$ towards neuron $N_j$. The CPC process is in charge to deliver the computation authorization, according to the following algorithm:\\
{\small
if the thread associated to $EC_j$ is already active\\
\indent then \{ $EC_j$ gets priority status [for further computation];\\
\indent \quad \quad computation is delayed; \}\\
\indent else if $st_i = pt_p$ then computation is authorized;\\
\indent \quad  else if $nbth_p = 0$ then\\
\indent \quad \quad \quad if $(\forall m) [st_i\leq et_m$ or $et_m\leq 0]$\\
\indent \quad \quad \quad \quad then computation is authorized;\\
\indent \quad \quad \quad \quad else if local deadlock is detected then\\
\indent \quad \quad \quad \quad \quad if $st_i \leq st_l(next\ event\ top\ of\ CPC\ queue)$\\
\indent \quad \quad \quad \quad \quad \quad then computation is authorized;\\
\indent \quad \quad else computation is delayed;\\
}
\noindent with the following condition for local deadlock detection:\\
{\small
\indent if $et_p < st_i$ and $(\forall m\neq p) [st_i\leq et_m$ or $et_m\leq 0]$\\
}
If the computation is authorized, the CPC updates both the local processing time: \quad {\small $pt_p \leftarrow st_i$}\\  and the number of locally active threads $nbth_p^{++}$. Each time the CMC queue becomes empty, the local processing time is changed to its opposite $pt_p \leftarrow - pt_p$.\\

The present algorithm authorizes the computation of only one event at a time by a given neuron $N_j$ (the computation is delayed if the thread of $EC_j$ is active) and regulates the computations, via the variable $pt_p$, according to the whole network advancement state, known by way of the clock $Clk(p)$ (controls on all the $et_m$). Once again, we avoid communication overhead, even if computation is delayed for a moment, due to an obsolete clock, since the problem will be solved by further reception of messages coming from other processors.\\

The {\em result collector} scans the active threads, first for an EC with priority status (if relevant) or in a loop on the number of the currently active threads. If a neuron $N_j$ computation of an event is over (i.e. thread ended), then the number of active threads is decremented $nbth_p^{--}$. The result of the computation is either null or a new outgoing spike event $[N_j, st_j, crt]$ that the {\em result collector} inserts in the CPC queue. If the CPC queue was previously empty, then the emission time $et_p$ takes back a positive value: {\small $et_p \leftarrow |et_p|$}.\\

Moreover, the computation of an event for a neuron $N_j$ induces the certification of old events $[N_j, st_j, crt]$ still present in the CPC queue. $crt \leftarrow$ ``true'' each time $st_j$ is less or equal to the currently processed $st_i$ plus the minimal delay $d_{j}^{min} = min_{i}(d_{ij})$ of neuron $N_j$.


\section{Conclusion}
\label{concl}

We have designed a framework dedicated to event-driven simulation of very large neural networks of biologically plausile spiking neurons. The DAMNED simulator is based on two levels of parallelism: At a coarse grain level, the SNN is distributed on several processors; At a fine grain level, local computations of neurons are multithreaded on each processor. Since local clock updates, based on event causality, are managed via spike events message passing, both time-consuming synchronization barrier and centralized farmer processor can be avoided.

Presently, the simulator has been successfully tested on a toy SNN, with a basic model of spiking neuron. Further work will include implementation of large heterogeneous SNNs. Time measurements and speed-up evaluations will be performed both on workstation clusters and on parallel computers (e.g. at IN2P3 and C3I computation centers).


{\small
\bibliographystyle{unsrt}
\bibliography{damned}

\begin{thebibliography}{10}

\bibitem{Sin99}
W.~Singer.
\newblock Neural synchrony: A versatile code for the definition of relations?
\newblock {\em Neuron}, 24:49--65, 1999.

\bibitem{TBF01}
C.~Tallon-Baudry, O.~Bertrand, and C.~Fischer.
\newblock Oscillatory synchrony between human extrastriate areas during visual
  short-term memory maintenance.
\newblock {\em J. Neuroscience}, 21:1--5, 2001.

\bibitem{IGE04}
E.M. Izhikevich, J.A. Gally, and G.M. Edelman.
\newblock Spike-timing dynamics of neuronal groups.
\newblock {\em Cerebral Cortex}, 14:933--944, 2004.

\bibitem{MeuPau05}
D.~Meunier and H.~Paugam-Moisy.
\newblock Inhibition and spike-time-dependent plasticity govern the formation
  and disruption of a distributed synchronized neural assembly.
\newblock (submitted), 2005.

\bibitem{Maa97}
W.~Maass.
\newblock Networks of spiking neurons: The third generation of neural network
  models.
\newblock {\em Neural Networks}, 10(9):1659--1671, 1997.

\bibitem{HopBro01}
J.J. Hopfield and C.D. Brody.
\newblock What is a moment? transient synchrony as a collective mechanism for
  spatiotemporal integration.
\newblock {\em Proc. Natl. Acad. Sci.}, 98(3):1282--1287, 2001.

\bibitem{LNM05}
R.~Legenstein, C.~Naeger, and W.~Maass.
\newblock What can a neuron learn with spike-time-dependent plasticity?
\newblock {\em Neural Computation}, 17(11):2337--2382, 2005.

\bibitem{SimSga05}
J.~Sima and J.~Sgall.
\newblock On the nonlearnability of a single spiking neuron.
\newblock {\em Neural Computation}, 17(12):2635--2647, 2005.

\bibitem{DGVT99}
A.~Delorme, J.~Gautrais, R.~Van~Rullen, and S.~Thorpe.
\newblock Spike{NET}: A simulator for modeling large networks of integrate and
  fire neurons.
\newblock {\em Neurocomputing}, 26-27:989--996, 1999.

\bibitem{BB98}
J.M. Bower and D.~Beeman.
\newblock {\em The Book of {GENESIS}: Exploring Realistic Neural Models with
  the GEneral SImulation System}.
\newblock Springer, 1998.
\newblock 2nd edition.

\bibitem{HC97}
M.L. Hines and N.T. Carnevale.
\newblock The {NEURON} simulation environment.
\newblock {\em Neural Computation}, 9:1179--1209, 1997.

\bibitem{Wat94}
L.~Watts.
\newblock Event-driven simulation of networks of spiking neurons.
\newblock In J.~D. Cowan, G.~Tesauro, and J.~Alspector, editors, {\em Advances
  in Neural Information Processing System}, volume~6, pages 927--934. MIT
  Press, 1994.

\bibitem{MDG00}
M.~Mattia and P.~Del~Giudice.
\newblock Efficient event-driven simulation of large networks of spiking
  neurons and dynamical synapses.
\newblock {\em Neural Computation}, 12:2305--2329, 2000.

\bibitem{Mak03}
T.~Makino.
\newblock A discrete event neural network simulator for general neuron model.
\newblock {\em Neural Computation and Applic.}, 11(2):210--223, 2003.

\bibitem{RocMar03}
O.~Rochel and D.~Martinez.
\newblock An event-driven framework for the simulation of networks of spiking
  neurons.
\newblock In {\em ESANN'03, European Symposium on Artificial Neural Network},
  pages 295--300, 2003.

\bibitem{RGF03}
J.~Reutimann, M.~Giugliano, and S.~Fusi.
\newblock Event-driven simulation of spiking neurons with stochastic dynamics.
\newblock {\em Neural Computation}, 15(4):811--830, 2003.

\bibitem{BAV99}
Y.~Boniface, F.~Alexandre, and S.~Vialle.
\newblock A library to implement neural networks on {MIMD} machines.
\newblock In {\em Proc. of Euro-Par}, pages 935--938, 1999.

\bibitem{EPPU02}
P.A. Est\'evez, H.~Paugam-Moisy, D.~Puzenat, and M.~Ugarte.
\newblock A scalable parallel algorithm for training a hierarchical mixture of
  neural networks.
\newblock {\em Parallel Computing}, 28:861--891, 2002.

\bibitem{JSRMK97}
A.~Jahnke, T.~Schoneauer, U.~Roth, K.~Mohraz, and H.~Klar.
\newblock Simulation of spiking neural networks on different hardware
  platforms.
\newblock In {\em ICANN'1997, Int. Conf. on Artificial Neural Networks}, pages
  1187--1192, 1997.

\bibitem{Sei04}
U.~Seiffert.
\newblock Artificial neural networks on massively parallel computer hardware.
\newblock {\em Neurocomputing}, 57:135--150, 2004.

\bibitem{HGGMRK05}
H.~H. Hellmich, M.~Geike, P.~Griep, M.~Rafanelli, and H.~Klar.
\newblock Emulation engine for spiking neurons and adaptive synaptic weights.
\newblock In {\em IJCNN'2005, Int. Joint Conf. on Neural Networks}, pages
  3261--3266. IEEE-INNS, 2005.

\bibitem{Fer95}
A.~Fersha.
\newblock Parallel and distributed simultation of discret event systems.
\newblock In A.~Y. Zomaya, editor, {\em Parallel and Distributed Computing
  Handbook}. McGraw-Hill, 1995.

\bibitem{GA98}
C.~Grassmann and J.~K. Anlauf.
\newblock Distributed, event-driven simulation of spiking neural networks.
\newblock In {\em NC'98, International ICSC/IFAC Symposium on Neural
  Computation}, pages 100--105. ICSC Academic Press, 1998.

\bibitem{PSWH01}
R.~Preis, K.~Salzwedel, C.~Wolff, and G.~Hartmann.
\newblock Efficient parallel simulation of pulse-coded neural networks (pcnn).
\newblock In {\em PDPTA'2001, International Conference on Parallel and
  Distributed Processing Techniques and Applications}, 2001.

\bibitem{GSW02}
C.~Grassmann, T.~Schoenauer, and C.~Wolff.
\newblock Pcnn neurocomputeurs - event driven and parallel architectures.
\newblock In {\em ESANN'02, European Symposium on Artificial Neural Nrtwork},
  pages 331--336, 2002.

\bibitem{Pau95}
H.~Paugam-Moisy.
\newblock Multiprocessor simulation of neural networks.
\newblock In M.~Arbib, editor, {\em The Handbook of Brain Theory and Neural
  Networks}, pages 605--608. MIT Press, 1995.

\bibitem{GK02}
W.~Gerstner and W.~Kistler.
\newblock {\em Spiking Neuron Models: Single Neurons, Populations, Plasticity}.
\newblock Cambridge University Press, 2002.

\bibitem{MB99}
W.~Maass and C.M. Bishop, editors.
\newblock {\em Pulsed Neural Networks}.
\newblock MIT Press, 1999.

\end{thebibliography}
}



\end{document}